\documentclass[sigconf,nonacm]{acmart}

\usepackage[table]{xcolor}
\usepackage{graphicx}
\usepackage{subcaption}
\usepackage{multirow}
\usepackage{amsmath}
\usepackage{amsfonts}
\usepackage{amsfonts}
\usepackage{dutchcal}
\AtBeginDocument{%
  }

\setcopyright{acmlicensed}
\copyrightyear{2018}
\acmYear{2018}
\acmDOI{XXXXXXX.XXXXXXX}
\acmConference[Conference acronym 'XX]{Make sure to enter the correct
  conference title from your rights confirmation email}{June 03--05,
  2018}{Woodstock, NY}
\acmISBN{978-1-4503-XXXX-X/2018/06}

\begin{document}

\title{Robust Adversarial Reinforcement Learning with Risk Sensitivity and Critic Consistency Regularization}

\author{Jiaxi Wu}
\email{jx-wu24@mails.tsinghua.edu.cn}
\affiliation{%
  \institution{Shenzhen International Graduate School, Tsinghua University}
  \city{Shenzhen}
  \country{China}
}

\author{Tiantian Zhang}
\authornotemark[1]
\email{zhang.tt@sz.tsinghua.edu.cn}
\affiliation{%
  \institution{Shenzhen International Graduate School, Tsinghua University}
  \city{Shenzhen}
  \country{China}
}

\author{Yuxing Wang}
\email{wyx20@mails.tsinghua.edu.cn}
\affiliation{%
  \institution{Shenzhen International Graduate School, Tsinghua University}
  \city{Shenzhen}
  \country{China}
}

\author{Yongzhe Chang}
\email{changyongzhe@sz.tsinghua.edu.cn}
\affiliation{%
  \institution{Shenzhen International Graduate School, Tsinghua University}
  \city{Shenzhen}
  \country{China}
}

\author{Xueqian Wang}
\email{wang.xq@sz.tsinghua.edu.cn}
\affiliation{%
  \institution{Shenzhen International Graduate School, Tsinghua University}
  \city{Shenzhen}
  \country{China}
}

\renewcommand{\shortauthors}{Wu et al.}
\begin{abstract}
  Reinforcement learning (RL) achieves strong performance in sequential decision-making but remains brittle under dynamic uncertainty and distributional shifts. Robust Adversarial Reinforcement Learning (RARL) improves robustness via worst-case perturbations, but existing approaches frequently suffer from unstable optimization and degraded value estimation. In particular, overly aggressive adversaries can drive the agent toward uninformative failure states, while adversarial perturbations amplify disagreement between double critics and introduce biased value targets. We propose a unified framework, RACER (\textbf{R}isk-sensitive robust \textbf{A}dversarial critic \textbf{C}onsist\textbf{E}ncy-regularized \textbf{R}einforcement learning), that revisits adversarial RL from a risk-sensitive perspective. First, we introduce a state-dependent adversarial objective that adaptively regulates perturbation strength, suppressing harmful disturbances while preserving informative exploration. Second, we propose critic consistency regularization to reduce disagreement between Q-value estimators and stabilize learning. Comprehensive experiments on challenging continuous control benchmarks demonstrate that RACER consistently improves performance, robustness, and training stability over strong robust RL baselines.
\end{abstract}

\maketitle

\section{Introduction}
Reinforcement learning (RL) has achieved significant success in sequential decision-making problems and has been widely applied across a range of domains, including robotic control \cite{dalal2021accelerating, H-GAP, luo2024serl}, autonomous driving \cite{osinski2020simulation, kiran2021deep, zhao2024survey} and adaptive healthcare systems \cite{coronato2020reinforcement, abdellatif2023reinforcement, lakhan2023drlbts}. However, its deployment in real-world environments remains fundamentally challenged by dynamic uncertainty and distributional shifts \cite{riahi2026distribution}. Policies trained in simulation often fail to generalize when exposed to perturbations, modeling errors, or unforeseen conditions, where even small deviations can accumulate into catastrophic failures \cite{ju2022transferring}. To address these challenges, robust RL has emerged as a critical research direction, aiming to learn policies that maintain reliable performance under uncertainty \cite{morimoto2005robust, wang2021online, moos2022robust}. Among existing approaches, Robust Adversarial Reinforcement Learning (RARL) \cite{pinto2017robust} has become a dominant paradigm, formulating the problem as a two-player zero-sum Markov game \cite{littman1994markov, perolat2015approximate} in which an adversary injects perturbations to minimize the protagonist’s return. By explicitly modeling worst-case disturbances, RARL provides a principled framework for improving robustness against environmental variations.

\begin{figure}[t]
\centering
\begin{subfigure}[t]{0.43\textwidth}
    \centering
    \includegraphics[width=\linewidth]{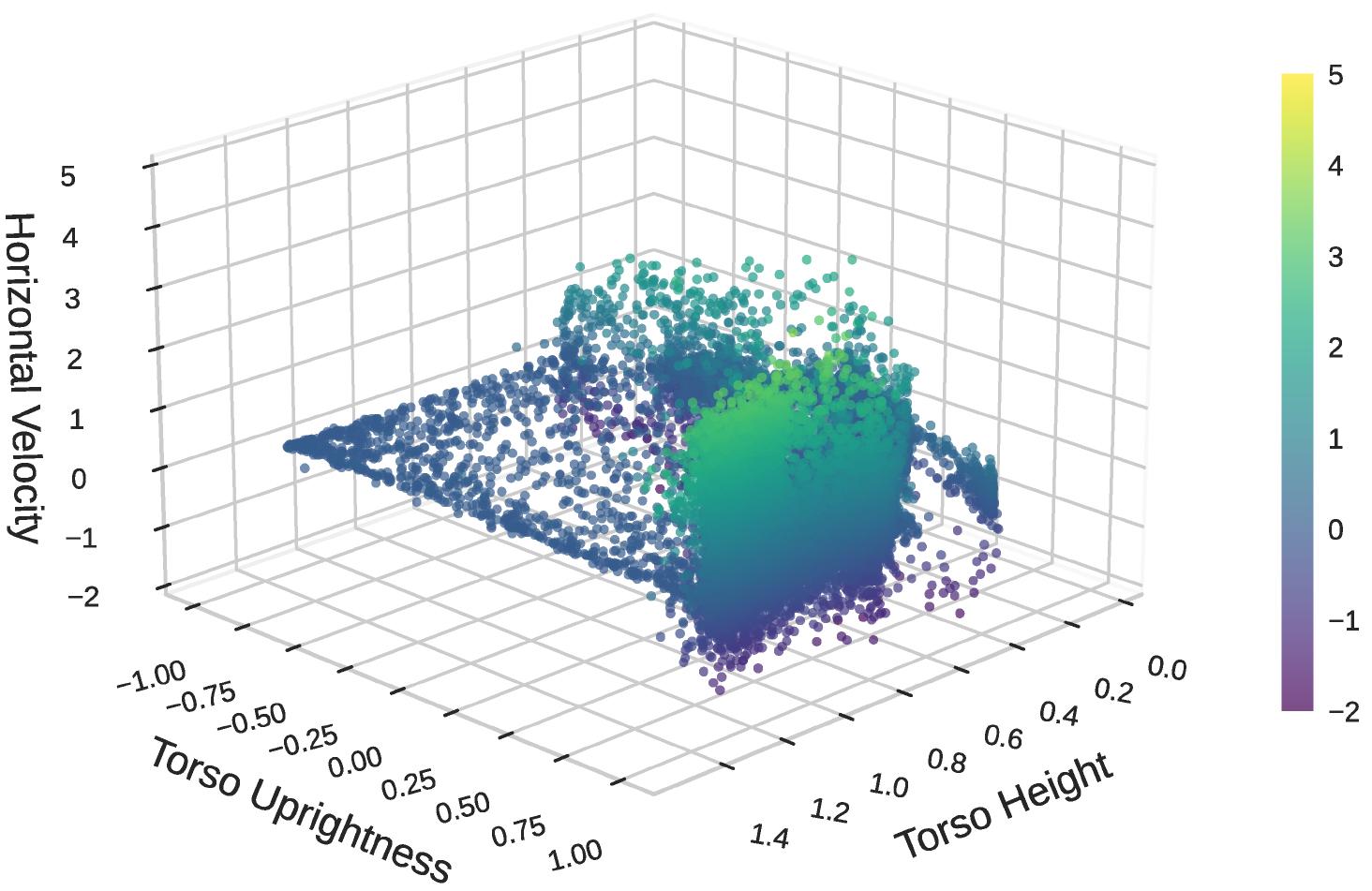}
    \caption{QARL}
\end{subfigure}
\hfill
\begin{subfigure}[t]{0.43\textwidth}
    \centering
    \includegraphics[width=\linewidth]{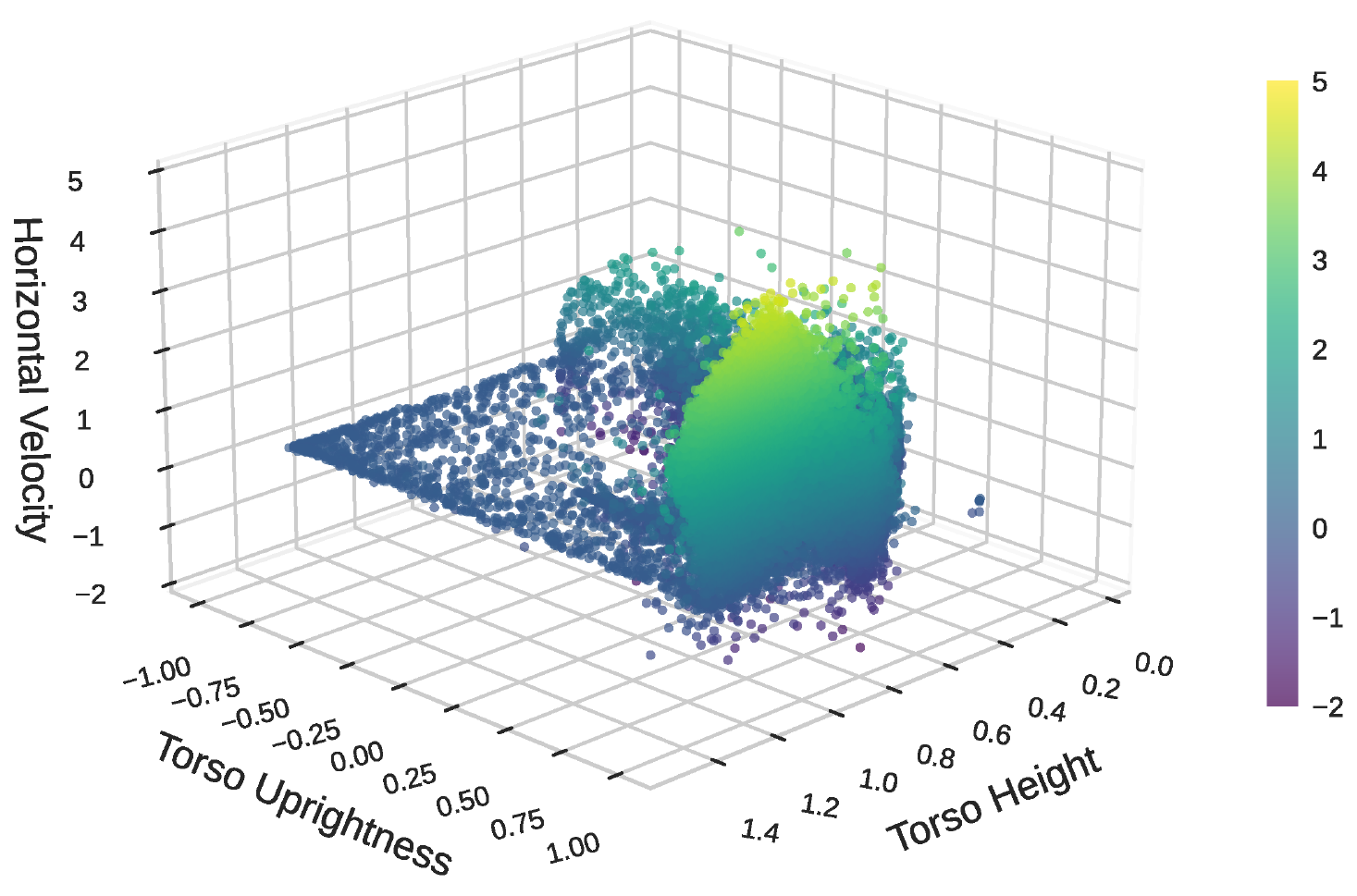}
    \caption{Risk-Sensitive QARL}
\end{subfigure}
\caption{Protagonist's state distributions during training in the MuJoCo locomotion task {\tt Walker-Run} under robust adversarial RL. Points are sampled from 50,000 randomly selected training steps, each representing the protagonist’s state, including torso height, uprightness, and horizontal velocity.}
\label{motivation}
\end{figure}

Despite its effectiveness, RARL suffers from several fundamental limitations. First, unconstrained adversarial perturbations can destabilize training by repeatedly driving the agent into extreme or unrecoverable states \cite{DBLP:conf/ijcnn/ShengZDKCZ22}. Since the adversary is optimized to produce worst-case disturbances, it may frequently drive the system into extreme or unrecoverable states, resulting in uninformative interactions and degraded learning signals for the protagonist. Consequently, the protagonist agent receives insufficient exposure to meaningful, near-optimal trajectories, hindering the acquisition of robust and stable policies. 
Empirically (as shown in Figure \ref{motivation}(a)), this manifests as a highly dispersed state distribution skewed toward failure-prone regions (low torso height, poor uprightness, near-zero velocity), indicating frequent collapse into unstable regimes.
Second, RARL suffers from a pronounced discrepancy between the two critics in value estimation, which undermines both the stability and accuracy of learning. In modern off-policy methods such as Soft Actor-Critic (SAC), double Q-networks are introduced to reduce overestimation by taking the minimum of two critics \cite{haarnoja2018soft, fujimoto2018addressing}. However, under adversarial perturbations, the two critics can diverge significantly for the same state–action pair. Since target computation relies on the minimum of the two critics, excessive disagreement introduces systematic underestimation bias and destabilizes policy optimization. 

To address these challenges, we revisit adversarial RL from a risk-sensitive perspective and propose a principled framework, RACER (\textbf{R}isk-sensitive robust \textbf{A}dversarial \textbf{C}onsist\textbf{E}ncy-regularized \textbf{R}einforcement learning), that jointly improves adversarial design and value estimation. Our key observation is that perturbations in high-risk states are often destructive rather than informative. Based on this insight, we introduce a risk-sensitive adversarial objective that incorporates a state-dependent risk function to regularize the adversary. This mechanism adaptively modulates adversarial strength, suppressing harmful perturbations in already unstable regions while maintaining sufficient exploration pressure in safer regimes. As a result, the adversary becomes informationally efficient, aligning its behavior with the learning needs of the protagonist rather than purely pursuing worst-case outcomes. 
As shown in Figure \ref{motivation}(b), our risk-sensitive formulation yields a more structured state distribution. Unlike the dispersed, failure-dominated distribution in Figure \ref{motivation}(a), the protagonist's states concentrate in a compact region corresponding to stable locomotion (higher torso height, better uprightness, consistent velocity), while retaining sufficient variability for exploration. This indicates that our method filters out uninformative perturbations while preserving those beneficial for policy improvement.
In addition, we introduce a critic consistency regularization framework that explicitly constrains disagreement between Q-value estimators in off-policy reinforcement learning. Concretely, we augment the standard double Q-networks objective with a regularization term that penalizes excessive divergence between critics, encouraging aligned value predictions during training. Our method interprets large critic discrepancies as a source of instability and suppresses them accordingly.

In summary, this work makes the following contributions: 
\begin{itemize}
    
    \item We propose a risk-sensitive adversarial formulation that adaptively modulates perturbation strength based on the protagonist’s state-dependent risk, improving adversarial training stability and robustness.
    
    \item We introduce critic consistency regularization, a scale-invariant objective that stabilizes double Q-learning under adversarial perturbations.
    
    \item We provide a unified framework, RACER, that bridges adversarial robustness and risk-sensitive learning, leading to improved the robustness of reinforcement learning across a range of challenging continuous control benchmarks.
    
\end{itemize}

\section{Related Work}

\subsection{Robust Adversarial Reinforcement Learning}

Robustness in RL has been widely studied to enable reliable decision-making under uncertainty \cite{moos2022robust}. A prominent line of work formulates the problem as a zero-sum game between a protagonist and an adversary, known as Robust Adversarial Reinforcement Learning \cite{pinto2017robust}. In this framework, the adversary perturbs actions, observations, or environment dynamics during training, exposing the agent to challenging scenarios and improving robustness under distributional shifts such as sim-to-real gaps. However, subsequent studies reveal inherent challenges in RARL. For instance, Zhang et al. \cite{NEURIPS2020_fb2e2032} show that RARL may suffer from instability and convergence difficulties even in simple linear-quadratic settings. MixedNE-LD \cite{DBLP:conf/nips/KamalarubanHHRS20} further points out that the underlying objective is generally non-convex–non-concave, making the joint optimization problem fundamentally hard. RARARL \cite{DBLP:conf/icra/PanSGC19} highlights that RARL lacks explicit modeling or optimization of risk and thus is unable to avoid catastrophic events. These limitations motivate a growing body of work that seeks to improve the stability, convergence, and robustness of adversarial RL.

\subsection{Holistic Optimization in Robust Adversarial Reinforcement Learning}

To address the limitations of RARL, a range of methods have been proposed from different perspectives, which can be broadly categorized based on the component of the optimization process they target:
(i) Optimization algorithms and training dynamics, which improve stability in adversarial games by enhancing optimization schemes or introducing explicit constraints. For example, DI-CARL \cite{DBLP:conf/aaai/ZhaiLDZWY22} provides stability guarantees via dissipation inequality constraints. Yu et al. \cite{yu2021robust} use Lagrangian duality and competitive mirror descent to improve stability and convergence, while MixedNE-LD \cite{DBLP:conf/nips/KamalarubanHHRS20} and Cen et al. \cite{DBLP:conf/nips/CenWC21} adopt stochastic gradient Langevin dynamics and extragradient methods, respectively.
(ii) Objective design and risk modeling, which incorporate risk or uncertainty into the objective to encourage safer policies. For instance, RARARL \cite{DBLP:conf/icra/PanSGC19} uses ensemble Q-value variance to model risk, promoting risk-averse behavior and reducing catastrophic failures.
(iii) Adversary scheduling and curricula, which regulate adversarial strength during training to improve stability. Methods such as CAT \cite{DBLP:conf/ijcnn/ShengZDKCZ22} and QARL \cite{DBLP:conf/iclr/ReddiT0CD24} apply curriculum learning to gradually increase difficulty, while A2P \cite{liu2024robust} introduces an adaptive adversarial coefficient to dynamically adjust perturbation strength.


\section{Preliminaries}

\subsection{Robust Adversarial Reinforcement Learning}

Robust Adversarial Reinforcement Learning models the problem as a two-player zero-sum Markov game \cite{littman1994markov, perolat2015approximate}, which is a special case of Markov games. In this setting, the protagonist and the adversary have strictly opposing objectives, where the adversary’s reward is defined as the negative of the protagonist’s reward. Formally, the two-player zero-sum Markov game can be represented as a tuple $\mathcal{M}=\langle \mathcal{S}, \mathcal{A}_p, \mathcal{A}_a, \mathcal{P}, \mathcal{R}, \gamma \rangle$, where $\mathcal{S}$ denotes the state space, and $\mathcal{A}_p$ and $\mathcal{A}_a$ are the action spaces of the protagonist and adversary, respectively. $\mathcal{P}:\mathcal{S}\times\mathcal{A}_p\times\mathcal{A}_a\times\mathcal{S}\rightarrow\mathbb{R}$ defines the transition dynamics, and $\mathcal{R}:\mathcal{S}\times\mathcal{A}_p\times\mathcal{A}_a\times\mathcal{S}\rightarrow\mathbb{R}$ denotes the reward function of both. The scalar $\gamma \in [0,1)$ is the discount factor.
Given a protagonist policy $\pi_p$ and an adversary policy $\pi_a$, the protagonist aims to maximize the expected discounted return, defined as:
\begin{equation}
J_{\pi_p,\pi_a}(s)=\mathbb{E}_{\pi_p, \pi_a, \mathcal{P}}\!\left[\sum\limits_{t=0}^{\infty} \gamma^{t} r(s_t, a_p^t, a_a^t, s_{t+1}) \right],
\end{equation}
where $a_p^t \sim \pi_{p}\left(\cdot \mid s_{t}\right), a_a^t \sim \pi_{a}\left(\cdot \mid s_{t}\right)$, and $s_{t+1} \sim \mathcal{P}(\cdot \mid s_{t}, a_a^t, a_p^t)$. Under the assumption of finite state and action spaces, this game is guaranteed to admit at least one Nash equilibrium. At equilibrium, the optimal joint policy $(\pi_p^*,\pi_a^*)$ satisfies the minimax theorem:
\begin{equation}
J_{\pi_p^\ast,\pi_a^\ast}(s)=\min_{\pi_a}\max_{\pi_p}J_{\pi_p,\pi_a}=\max_{\pi_p}\min_{\pi_a}J_{\pi_p,\pi_a}.
\end{equation}
This formulation highlights that the two-player zero-sum Markov game can be viewed as a max–min optimization problem, where the protagonist seeks to maximize the expected return while the adversary aims to minimize it by selecting worst-case perturbations \cite{littman1994markov, perolat2017learning, songcan}.

\subsection{Soft Actor-Critic}

Following prior work on robust adversarial RL \cite{pinto2017robust, DBLP:conf/nips/KamalarubanHHRS20, DBLP:conf/iclr/ReddiT0CD24}, we adopt SAC \cite{haarnoja2018soft} as the underlying RL algorithm in this paper. SAC is an off-policy actor–critic method that augments the standard return maximization objective with an entropy regularizer, encouraging persistent exploration and improved robustness.
Concretely, SAC learns a stochastic policy $\pi_{\phi}(a_t \mid s_t)$ and a soft Q-function $Q_{\theta}(s_t,a_t)$. The objective is to maximize the expected discounted return while promoting high-entropy policies. The Q-function is trained by minimizing the soft Bellman residual:
\begin{equation}
\mathcal{L}_{Q}(\theta) = \mathbb{E}_{(s_t,a_t,r_t,s_{t+1}) \sim \mathcal{D}}
\left[\Big(Q_\theta(s_t, a_t) - \big(r_t + \gamma V(s_{t+1})\big)\Big)^2\right],
\end{equation}
where the soft value function is defined as:
\begin{equation}
V(s_t) = \mathbb{E}{a_t \sim \pi_{\phi}} \big[ Q_{\theta}(s_t, a_t) - \alpha \log \pi_{\phi}(a_t \mid s_t) \big],
\end{equation}
and $\mathcal{D}$ denotes the replay buffer, $\alpha$ is the temperature parameter that controls the strength of entropy regularization. The policy is updated by minimizing
\begin{equation}
\mathcal{L}_{\pi}(\phi) = \mathbb{E}{s_t \sim \mathcal{D}}
\left[ \mathbb{E}{a_t \sim \pi_{\phi}} \big[ \alpha \log \pi_{\phi}(a_t \mid s_t) - Q_{\theta}(s_t, a_t) \big] \right],
\end{equation}
which trades off reward maximization against entropy maximization. These coupled updates yield a stable and sample-efficient learning procedure.

\section{Methodology}

In this section, we give a detailed description of the unified framework RACER. Specifically, we introduce a risk-sensitive adversarial formulation that regularizes the adversary with a state-dependent risk functional, enabling adaptive and informative perturbations instead of purely worst-case disturbances. In parallel, we incorporate a critic consistency regularization into SAC to suppress excessive disagreement between critics under adversarial perturbations. Together, these components improve both training stability and robustness, yielding a more reliable robust adversarial RL framework in uncertain environments.

\subsection{Risk-Sensitive Adversarial Formulation}



We revisit robust adversarial reinforcement learning from the perspective of risk-sensitive decision-making in stochastic games. Standard RARL formulates the interaction between the protagonist and the adversary as a two-player zero-sum Markov game, where the adversary is trained to induce worst-case perturbations by directly minimizing the protagonist’s return. While this formulation provides robustness guarantees under bounded disturbances, it implicitly assumes that the worst-case perturbation is always informative for policy improvement. However, this assumption is often violated in practice. Unconstrained adversaries tend to exploit vulnerabilities of the learning process rather than the policy itself, leading to degenerate training dynamics. In particular, excessively aggressive perturbations can drive the system into extreme regions of the state space, where the induced trajectories carry limited learning signals and hinder effective policy optimization \cite{DBLP:conf/ijcnn/ShengZDKCZ22}. To address this issue, we propose to reinterpret adversarial interaction as a risk-sensitive game, where the adversary is no longer purely worst-case driven but instead optimized under a state-dependent, regularized objective. 
Let the protagonist agent's state be $\mathbcal{s} = [\mathbcal{s}_1, \mathbcal{s}_2, \ldots, \mathbcal{s}_d] \in \mathbb{R}^d$. We construct $R(\mathbcal{s})$ as a \emph{composite} functional that aggregates multiple physically interpretable failure modes:
\begin{equation}
R(\mathbcal{s})=\beta_1 \times R_1(\mathbcal{s}_1)+\beta_2 \times R_2(\mathbcal{s}_2)+ \ldots + \beta_d \times R(\mathbcal{s}_d) = \sum_{i=1}^{d} \beta_i R_i(\mathbcal{s}_i),
\end{equation}
where $R_i(\mathbcal{s}_i)$ is used to describe an aspect of instability (e.g., height, velocity, and control-related risks), capturing failure modes such as falling or loss of balance of the protagonist. Depending on the semantics of the state variable, the risk function can be instantiated as either lower-bound ($\max(0, \epsilon_i - \mathbcal{s}_i)$) or upper-bound ($\max(0, \mathbcal{s}_i - \epsilon_i)$) violations. $\epsilon_i$ are the risk threshold for $\mathbcal{s}_i$. 
$\beta_i$ are non-negative and constrained weighting coefficients, which are optimized jointly with the policy parameters through gradient-based updates induced by the agent's objective.


\begin{figure}
 \includegraphics[width=0.34\textwidth]{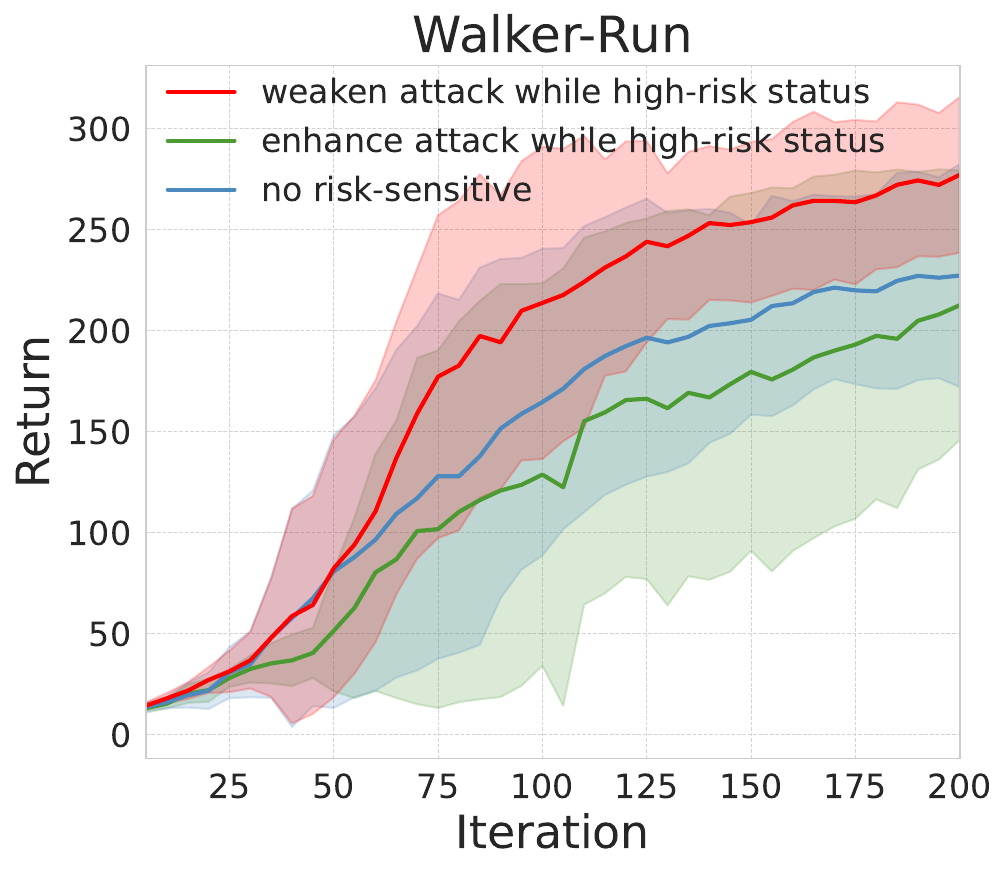}
  \caption{Learning curves on the MuJoCo locomotion task {\tt Walker-Run} comparing three risk-adaptive adversarial strategies. Shaded regions denote standard deviation across 5 random seeds.}
  \label{risk_change}
\end{figure}

If we use $r_p$ to denote the protagonist's reward and $r_a$ to denote the adversary's reward, then we can incorporate it into the adversary's objective as a soft constraint:
\begin{equation}
r_a=-\left(r_p + R(\mathbf{\mathbcal{s}}) \right).
\end{equation}
The key idea is that the risk term $R(\mathbcal{s})$ penalizes the adversary when the protagonist is already in a high-risk state, which allows our approach to induce a state-adaptive adversarial policy: in high-risk regions, the regularization term dominates, suppressing destructive perturbations; in low-risk regions, the adversary recovers its ability to generate strong and informative disturbances. Figure \ref{risk_change} illustrates this mechanism empirically on the Walker-Run benchmark: when the protagonist is in high-risk states, weakening the adversary's attack strength (\textcolor{red!50}{red curve}) yields substantially higher returns and lower variance compared to both enhancing attacks (\textcolor{green!40}{green curve}) and a risk-agnostic baseline (\textcolor{blue!50}{blue curve}). Notably, enhancing attacks during high-risk states severely destabilizes training and degrades asymptotic performance, confirming that unregulated adversarial pressure in fragile states can significantly destabilize learning. This adaptive behavior effectively reshapes the adversarial policy class from maximally destructive to informationally efficient, aligning the adversary's objective with the learning needs of the protagonist. As a result, the proposed formulation improves training stability while preserving sufficient adversarial pressure for robustness.


More broadly, our method can be viewed as an instance of risk-regularized minimax optimization, bridging robust reinforcement learning and risk-sensitive control. Despite this conceptual connection, our approach differs fundamentally from RARARL \cite{DBLP:conf/icra/PanSGC19}. Specifically, RARARL employs an ensemble of Q-networks and uses the variance of Q-value estimates as a proxy for risk, thereby discouraging actions with high uncertainty. In contrast, we model the protagonist’s risk explicitly in the state representation, enabling direct and structured access to risk information at the state level. By explicitly capturing the interaction between disturbance intensity and state-dependent risk, our framework provides a principled mechanism to mitigate the conservatism and training instability that commonly arise in adversarial reinforcement learning. Empirically, we observe significantly improved robustness and training stability.


\subsection{Critic Consistency Regularization}
\label{critic_method}

SAC employs a double-critic architecture, where two independent Q-functions $Q_{\phi_0}$ and $Q_{\phi_1}$, parameterized by $\phi_0$ and $\phi_1$, are used to estimate the state–action value for a given pair $(s,a)$. For notational convenience, we denote their outputs as $q_i=Q_{\phi_i}(s,a),i \in \{0,1\}$. The minimum of the two estimates is used to mitigate overestimation bias induced by function approximation. However, in robust adversarial reinforcement learning, the presence of adversarial perturbations significantly increases the uncertainty of Q-value estimation. This often leads to severe discrepancies between the two critics for the same input. In such cases, the minimum operator does not merely reduce overestimation, but systematically selects the lower estimate, effectively converting critic disagreement into a persistent underestimation bias \cite{fujimoto2018addressing}. This effect is exacerbated when the discrepancy between critics becomes large, which can hinder effective policy learning. To address this issue, we propose a critic consistency regularization method that explicitly penalizes excessive disagreement between the two critics. The key idea is to suppress adversarially induced divergence while preserving the conservative nature of double Q-networks, thereby improving both the stability and robustness of value estimation. We define a discrepancy measure which corresponds to the sample variance of two critics: 
\begin{equation}
\mathcal{V}(q_0,q_1)=\frac{(q_0-\mu)^2+(q_1-\mu)^2}{2}=\frac{(q_0-q_1)^2}{4}, \quad \mu=\frac{q_0+q_1}{2}.
\end{equation}
This formulation corresponds to a scaled pairwise squared difference and serves as a smooth and differentiable measure of critic disagreement. A key challenge is that the magnitude of Q-values varies significantly across tasks, making the regularization strength sensitive to reward scale. To address this, we introduce a normalization factor $S$:
\begin{equation}
S=E_{(s,a) \sim \mathcal{D}}[\frac{|q_0|+|q_1|}{2}],
\end{equation}
which is treated as a constant during backpropagation. $\mathcal{D}$ denotes the replay buffer, which is maintained separately for the protagonist and the adversary.
Using this normalization, we define the consistency regularization term:
\begin{equation}
\mathcal{L}_{cons}(q_0,q_1)=E_{(s,a) \sim \mathcal{D}}\frac{(q_0-q_1)^2}{4(S+\delta)},
\end{equation}
where $\delta>0$ is a small fixed constant for numerical stability, requiring no task-specific tuning. This normalization makes the regularizer approximately scale-invariant, allowing it to adapt to different Q-value magnitudes and reducing sensitivity to task-specific reward scales.
For both the protagonist and the adversary, we adopt independent critic networks and optimize them using the same objective. Specifically, for each agent $j \in \{p, a\}$, where $p$ and $a$ denote the protagonist and the adversary, the critic loss under the double critic setting is defined as:
\begin{equation}
\label{lambda_equation}
\mathcal{L}_{Q_j}=\sum_{k=0}^{1} \mathbb{E}_{\left(s, a_j, r_j, s^{\prime}\right) \sim \mathcal{D}_j}\left[\left(Q_{\phi_{j,k}}(s, a_j)-y_j\left(r_j, s^{\prime}\right)\right)^{2}\right]+\lambda \mathcal{L}_{\mathrm{cons}_{j}},
\end{equation}
where $\lambda$ is a consistency regularization coefficient and is the same across both agents. The target value is defined as:
\begin{equation}
y_j(r_j,s')=r_j+\gamma\left(\min _{k=0,1} Q_{\bar{\phi}_{j,k}}\left(s^{\prime}, a_j^{\prime}\right)-\alpha_j \log \pi_j\left(a_j^{\prime} \mid s^{\prime}\right)\right),
\end{equation}
where $a_j' \sim \pi_j(\cdot \mid s')$, and $\gamma \in [0,1)$ is the discount factor.




\begin{figure}[t]
\centering
\begin{subfigure}[t]{0.35\textwidth}
    \centering
    \includegraphics[width=\linewidth]{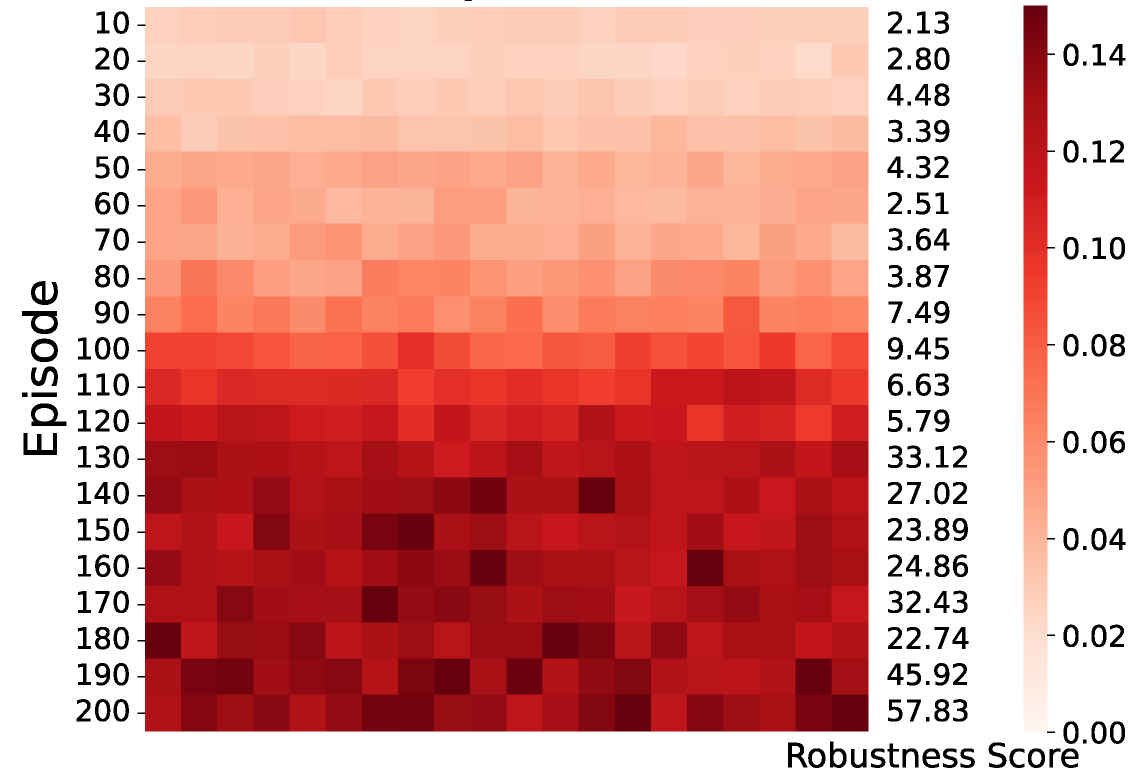}
    \caption{QARL}
\end{subfigure}
\hfill
\begin{subfigure}[t]{0.35\textwidth}
    \centering
    \includegraphics[width=\linewidth]{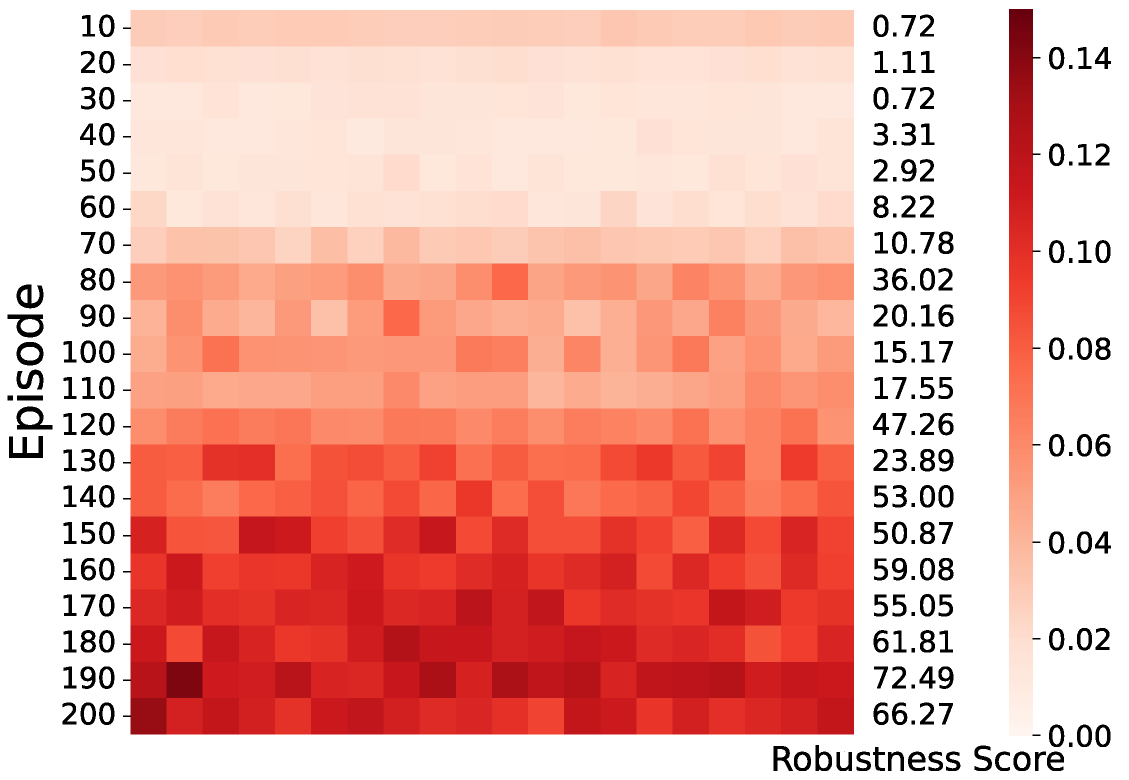}
    \caption{RACER}
\end{subfigure}

\caption{Evolution of critic disagreement and robustness during training for RACER and QARL.}
\label{critic_consistency}
\end{figure}

Unlike prior works that leverage ensemble variance for exploration or uncertainty estimation \cite{kalweit2017uncertainty, chen2017ucb}, our method explicitly reduces critic disagreement to counteract adversarially induced divergence. The normalization further distinguishes our approach by making the penalty adaptive to value scale. A potential concern is that the proposed consistency regularization may collapse the two critic networks into identical functions, thus diminishing the benefit of double Q-networks. Importantly, our method does not enforce exact equality between the two critics. Instead, it penalizes excessive discrepancy relative to the value scale, allowing meaningful diversity to be preserved. Moreover, the minimum operator used in target computation remains unchanged, ensuring that the conservative bias of double Q-networks is retained. 

To substantiate this argument, we compare QARL and RACER by quantifying the discrepancy between the two SAC critics over state–action samples. Specifically, for each training episode, we compute the absolute difference between the Q-values produced by the two critics for all $(s,a)$ pairs within the batch size and report their mean, yielding an episode-level signal that reflects critic consistency. As shown in Figure \ref{critic_consistency}, we visualize this signal as heatmaps over the course of training, where each row corresponds to an episode and each column indexes a fixed sampled position within the batch. Darker colors indicate larger disagreement between the critics. The corresponding robustness evaluation score at each episode is reported on the right side of each panel. The results suggest that RACER maintains a more controlled and stable level of critic disagreement compared to QARL while preserving sufficient diversity between critics, and this behavior is accompanied by consistently higher robustness scores throughout training.


\section{Experiments}

To comprehensively evaluate our method, we conduct experiments addressing the following research questions:  

\begin{itemize}
    \item[Q1] Can RACER achieve more robust adversarial reinforcement learning than existing methods under environmental distribution shifts and adversarial perturbations?
    
    \item[Q2] Which components of RACER are most responsible for its performance gains?
    
    \item[Q3] How does the consistency regularization coefficient $\lambda$ influence overall performance?  
    
    \item[Q4] Can RACER be seamlessly integrated into other existing robust adversarial RL algorithms?
\end{itemize}

\subsection{Experimental Setup}

\paragraph{Environments.} 
We consider a diverse set of MuJoCo control problems \cite{todorov2012mujoco} from the DeepMind Control Suite \cite{tunyasuvunakool2020dm_control}:
{\tt Walker-Run}, {\tt Hopper-Stand}, {\tt Cheetah-Run}, and {\tt Cartpole-Balance}.
Following the adversarial training protocols established in prior works \cite{pinto2017robust,DBLP:conf/nips/KamalarubanHHRS20,DBLP:conf/iclr/ReddiT0CD24}, in each environment, we introduce specific high-intensity disturbances through adversarial actions and force magnitudes from the adversary to elicit robust protagonist agent behavior. More details be found in \emph{Appendix \ref{environment_settings}}.

\paragraph{Baselines.} We benchmark our method against the classical robust adversarial RL baseline RARL \cite{pinto2017robust} and its recent extensions: (i) MixedNE-LD \cite{DBLP:conf/nips/KamalarubanHHRS20}, which employs Langevin dynamics to escape local equilibria, (ii) CAT \cite{DBLP:conf/ijcnn/ShengZDKCZ22}, which anneals adversary strength through a manually-crafted curriculum, (iii) the current state-of-the-art QARL \cite{DBLP:conf/iclr/ReddiT0CD24}, an algorithm that couples entropy regularization with quantal response equilibrium while modulating adversary rationality via temperature-scheduled curriculum learning. Standard SAC \cite{haarnoja2018soft} serves as the non-adversarial backbone for reference.
Unless specified otherwise, RACER implementations in the following build upon the QARL framework. For each comparison method, we conduct training with five random seeds $\{0,1,2,3,4\}$ across all environments, performing $200$ alternating iterations per seed. All reported results in our experiments represent the average performance across these seeds.
Complete hyperparameter configurations for all algorithms are provided in \emph{Appendix \ref{hyperparameter_settings}}.

\paragraph{Hardware and Software.} All experiments were run on a workstation equipped with an NVIDIA TITAN. Detailed hardware and software conditions can be found in \emph{Appendix \ref{hardware}}.

\subsection{Performance Comparison}

\begin{figure*}[!t]
\centering
\includegraphics[width=0.95\linewidth]{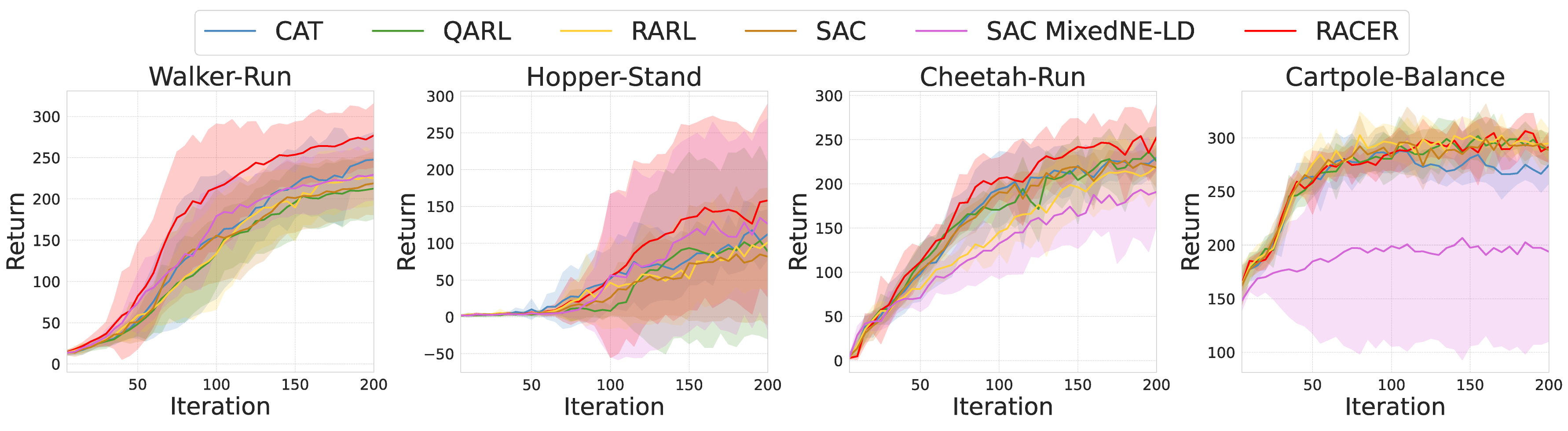} 
\caption{Training curves comparing RACER against five baselines across four MuJoCo control and locomotion tasks. Curves show mean robustness performance over 5 random seeds with shaded regions indicating standard deviation.}
\label{training_process_robustness}
\end{figure*}

\begin{figure*}[!t]
\centering
\includegraphics[width=0.98\linewidth]{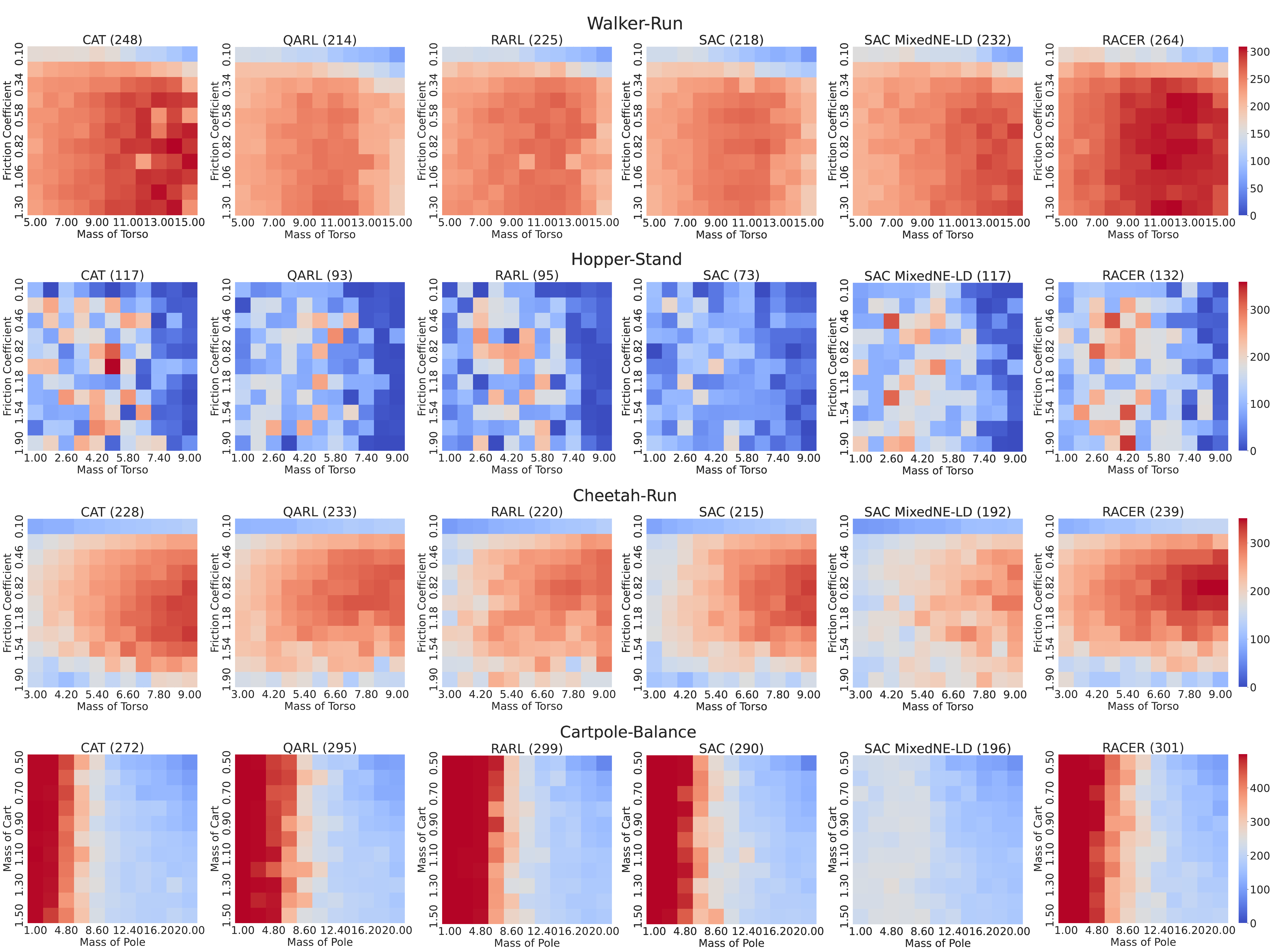} 
\caption{Final robustness evaluation results of RACER and baseline methods across four MuJoCo control and locomotion tasks, evaluated at the end of training. Each heatmap illustrates the performance achieved for different environmental properties, which are described on the x-y axes.}
\label{heatmap}
\end{figure*}

\begin{table*}[!t]
\caption{Final adversarial performance comparing RACER against four baselines across four MuJoCo control and locomotion tasks, evaluated at the end of training against an adversary trained against the frozen trained protagonist, reporting mean final robustness performance $\pm$ standard deviation over 5 random seeds. 
The record corresponding to SAC does not have adversarial results because it lacks an adversarial agent. Therefore, its scores are the final results obtained through independent training of the protagonist agent. The optimal performance is marked with \textcolor{blue!40}{blue backgrounds}.}
\label{vs_worst_adversary}
\centering
\resizebox{0.78\hsize}{!}{
\setlength{\tabcolsep}{2.6mm}
\begin{tabular}{c|cccc}
  \toprule
  Method           & Walker-Run        & Hopper-Stand                     & Cheetah-Run      & Cartpole-Balance          \\
  \midrule
  SAC   & $260.905 \pm 55.821$ & $119.141 \pm 159.773$ & $275.005 \pm 61.997$ & $498.480 \pm 0.282$ \\
  \midrule
  RARL  & $257.679 \pm 50.298$ & $153.350 \pm 111.529$ & $268.101 \pm 52.009$ & $497.802 \pm 1.226$ \\
  SAC MixedNE-LD    & $233.273 \pm 41.541$ & $135.032 \pm 215.379$ & $227.447 \pm 54.054$ & $238.004 \pm 179.891$ \\
  CAT   & $229.088 \pm 23.850$ & $143.994 \pm 70.863$ & $192.526 \pm 14.081$ & $301.107 \pm 21.483$ \\
  QARL  & $253.536 \pm 46.556$ & $145.296 \pm 196.174$ & $292.861 \pm 42.095$ & $498.260 \pm 0.565$ \\
  RACER (ours)   & \cellcolor{blue!15}$299.944 \pm 46.721$ & \cellcolor{blue!15}$190.073 \pm 190.661$ & \cellcolor{blue!15}$310.067 \pm 26.959$ & \cellcolor{blue!15}$498.512 \pm 0.318$ \\
  \bottomrule
\end{tabular}
}
\end{table*}

\begin{figure*}[!t]
\centering
\includegraphics[width=0.95\linewidth]{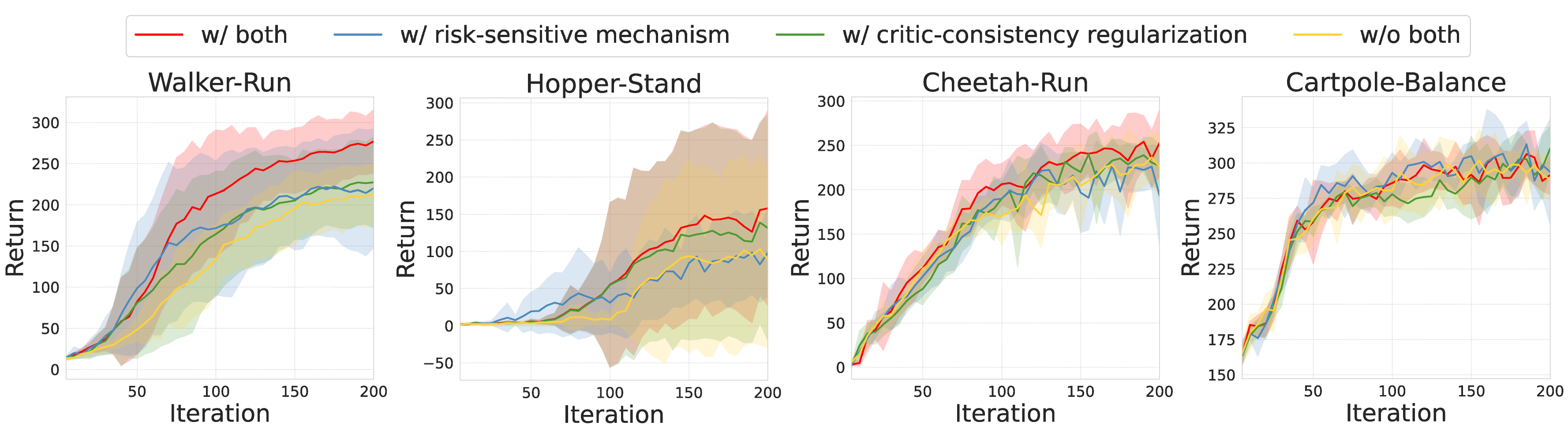} 
\caption{Ablation study on four algorithmic variants across four MuJoCo control and locomotion tasks. 
Curves show mean robustness performance over 5 random seeds with shaded regions indicating standard deviation.}
\label{ablation}
\end{figure*}

\paragraph{Training Robustness and Stability.} To better assess the performance of the model, we adopt a periodic robustness evaluation protocol. Specifically, for each environment, we measure the protagonist’s return by sweeping over a predefined range of environmental parameter perturbations (e.g., variations in mass and friction coefficients) in the absence of an adversary every five training iterations. We then report the average return across all parameter configurations as the \emph{robustness metric} to characterize the robustness performance of the protagonist at the current stage.

Figure \ref{training_process_robustness} presents the robustness evaluation of RACER and baseline methods across four continuous control benchmark tasks. RACER consistently achieves superior robustness scores in all environments, demonstrating not only faster convergence and more stable training dynamics but also strong generalization capabilities. During the early stage of training (the first 50–80 iterations), RACER exhibits a noticeably faster improvement in robustness, suggesting more stable policy learning and adaptation under adversarial perturbations. At the convergence period, RACER attains higher robustness ceilings on multiple tasks while maintaining narrower variance bands (shaded regions), indicating improved stability and generalization across different random seeds. Meanwhile, throughout the entire training process, RACER maintains a more stable training curve. Notably, on the more challenging Hopper-Stand task, although all methods exhibit relatively large variance, RACER still achieves better average performance, further highlighting its robustness in complex control scenarios.

\paragraph{Final Robustness.} To rigorously evaluate the final robustness of RACER, we perform a comprehensive analysis of its final-stage performance. Figure \ref{heatmap} reports the episode returns of the protagonist under systematically perturbed dynamics during the robustness evaluation phase. These perturbations span environment-specific parameters, including key physical properties such as friction coefficients, body masses, and joint damping, capturing realistic sources of uncertainty encountered in deployment. As shown in Figure \ref{heatmap}, RACER consistently maintains strong performance across a wide range of challenging conditions, exhibiting superior adaptability and robustness to variations in underlying dynamics.

\paragraph{Worst-Case Performance.} Beyond the comprehensive robustness analyses under diverse environmental conditions, we further adopt {\em final adversarial performance} through minimax perspective to examine RACER’s resilience against stronger, previously unseen adversaries. Specifically, after training converges, we fix the learned protagonist and continue optimizing an adversary with the explicit objective of degrading the protagonist’s performance. The final evaluation is then conducted between the frozen protagonist and this newly optimized worst-case adversary. As quantitatively reported in Table~\ref{vs_worst_adversary}, RACER consistently outperforms all baseline methods across different environments when confronted with these more challenging adversaries, while exhibiting low sensitivity to initialization, as evidenced by reduced performance variance. These results underscore RACER’s robustness under extreme adversarial settings and demonstrate its strong generalization to perturbations beyond those encountered during training.

\subsection{Ablation Study}

\begin{table*}[!t]
\caption{Sensitivity analysis of the consistency regularization coefficient $\lambda$ across four MuJoCo control and locomotion tasks, reporting mean final robustness performance $\pm$ standard deviation over 5 random seeds. The optimal performance is marked with \textcolor{blue!40}{blue backgrounds}.}
\label{hyperparameter}
\centering
\resizebox{0.74\hsize}{!}{
\setlength{\tabcolsep}{2.6mm}
\begin{tabular}{c|cccc}
  \toprule
  $\lambda$           & Walker-Run        & Hopper-Stand                     & Cheetah-Run      & Cartpole-Balance          \\
  \midrule
  0 (QARL) & $213.593 \pm 47.806$ & $92.952 \pm 131.105$ & $232.889 \pm 34.863$ & $295.086 \pm 28.063$\\
  \midrule
  $1 \times 10^{-2}$ & $216.220 \pm 48.628$ & $122.039 \pm 33.517$ & $240.187 \pm 26.429$ & $298.023 \pm 36.310$\\
  $1 \times 10^{-3}$ & $205.180 \pm 56.150$ & $88.113 \pm 57.543$ & $220.614 \pm 20.523$ & $285.467 \pm 16.261$\\
  $1 \times 10^{-4}$ & \cellcolor{blue!15}$263.520 \pm 46.196$ & \cellcolor{blue!15}$131.786 \pm 143.546$ & \cellcolor{blue!15}$240.770 \pm 20.870$ & \cellcolor{blue!15}$301.454 \pm 35.707$\\
  $1 \times 10^{-5}$ & $238.521 \pm 55.458$ & $67.013 \pm 62.534$ & $231.933 \pm 33.492$ & $293.535 \pm 23.122$\\
  \bottomrule
\end{tabular}
}
\end{table*}

We conduct controlled single-variable ablation studies within robust adversarial reinforcement learning settings to evaluate RACER against three algorithmic variants to systematically assess the contribution of its individual components.
(i) {\em w/ risk-sensitive mechanism}, which retains the risk-sensitive objective while disabling critic-consistency regularization;
(ii) {\em w/ critic-consistency regularization}, which preserves the consistency constraint on the critic while removing the risk-sensitive formulation;
(iii) {\em w/o both}, which removes both the risk-sensitive mechanism and the critic-consistency regularization, reducing the method to a standard baseline (QARL) formulation.

The results presented in Figure \ref{ablation} indicate that removing any individual component from RACER leads to varying degrees of performance degradation, with the most pronounced drop observed when both components are simultaneously disabled. This suggests that the two mechanisms are complementary and jointly contribute to improved robustness and stability under adversarial perturbations. 
The risk-sensitive mechanism mainly improves early-stage learning, leading to faster convergence and better exploration under adversarial perturbations. In contrast, critic-consistency regularization primarily enhances training stability, reflected in reduced variance and smoother learning dynamics. Moreover, the performance gap between the full model and its ablated variants increases over time, suggesting compounding benefits of both components. Overall, the “w/o both” variant consistently underperforms, highlighting the necessity of jointly modeling risk sensitivity and critic stability in adversarial reinforcement learning.

\subsection{Hyperparameter Analysis}

We conduct a systematic study to understand the impact of the consistency regularization coefficient $\lambda$ in Eq. (\ref{lambda_equation}) on learning stability and performance under adversarial perturbations. Recall that $\lambda$ controls the strength of the penalty on critic disagreement, thereby governing the trade-off between consistency and diversity.
We vary $\lambda$ over a wide range, from $0$ to $1 \times 10^{-2}$, and evaluate performance across multiple environments. When $\lambda = 0$, the method reduces to standard double critic training. 

Table \ref{hyperparameter} presents the final robustness evaluation results for different values of $\lambda$. Overall, the effect of $\lambda$ is not strictly monotonic across tasks. Compared to $\lambda = 0$, introducing a small amount of regularization can improve robustness in several environments, but performance may fluctuate at intermediate values (e.g., $\lambda = 10^{-3}$). Notably, when $\lambda = 1 \times 10^{-4}$, the method achieves the best or near-best performance across all tasks, indicating that a moderate level of regularization provides the most consistent benefit. This suggests that properly balancing critic agreement helps stabilize Q-value estimation and improves policy learning. However, when $\lambda$ deviates from this optimal range, either becoming too small or too large, performance may degrade. Importantly, even at relatively large $\lambda$, we do not observe catastrophic failure or collapse of the critics. These results indicate that $\lambda$ acts as a sensitive but stable control knob for critic regularization, with moderate values (around $10^{-4}$) providing the best trade-off without requiring extensive environment-specific tuning.

\subsection{Scalability Analysis}

\begin{figure}
  \includegraphics[width=0.85\linewidth]{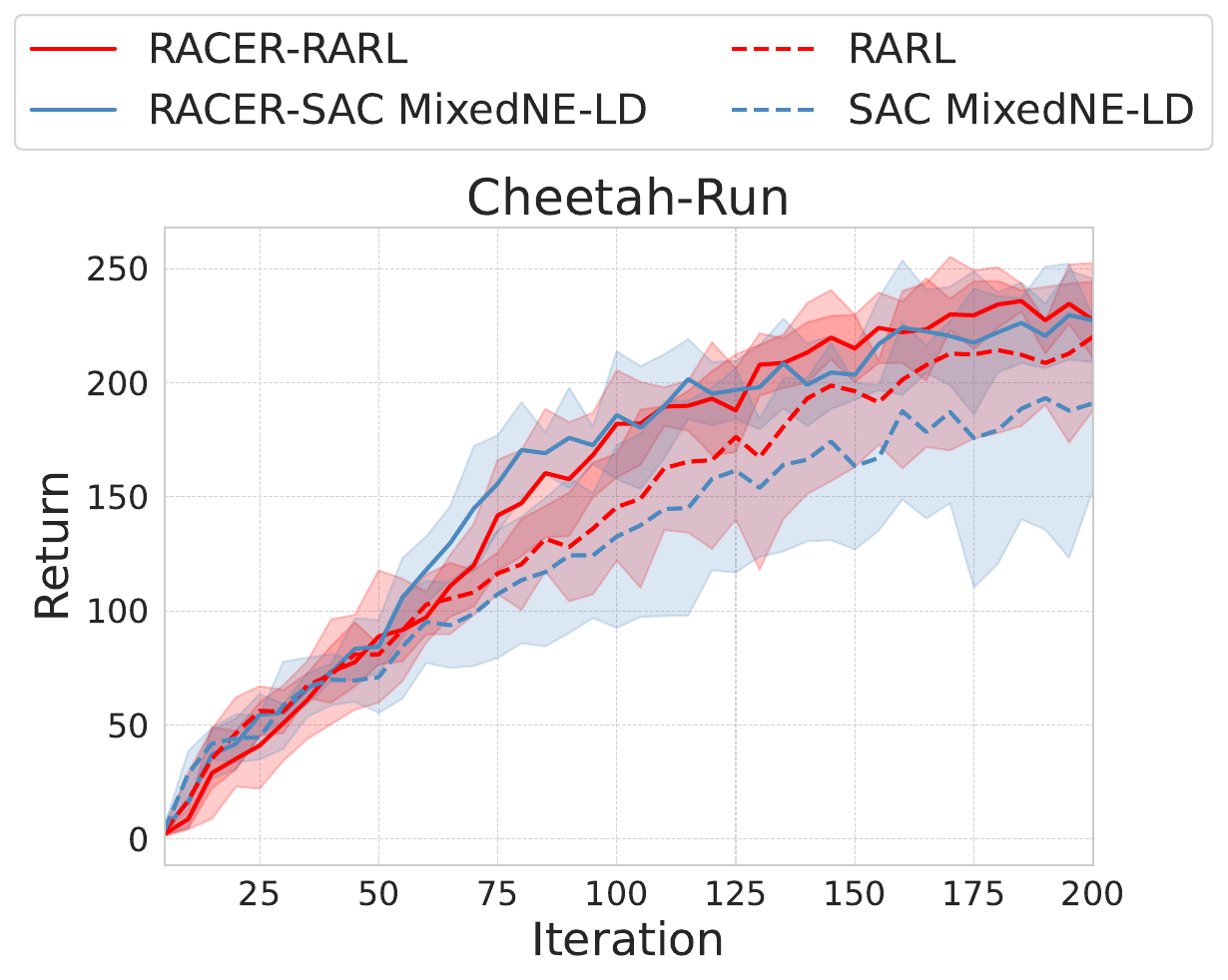}
  \caption{Learning curves of the MuJoCo locomotion task {\tt Cheetah-Run}, comparing the performance of RARL and MixedNE-LD before (dashed lines) and after (solid lines) the integration of the RACER framework. Shaded regions denote standard deviation across 5 random seeds.}
  \label{scalability}
\end{figure}

To evaluate the scalability of our approach, we further investigate whether the proposed RACER framework can generalize beyond its original instantiation on QARL. Actually, the design of RACER is largely decoupled from any specific adversarial update scheme and instead builds upon the underlying SAC optimization. This property enables us to seamlessly extend RACER to other representative robust adversarial reinforcement learning methods that also rely on SAC, including RARL and SAC MixedNE-LD.

Concretely, we construct two variants, RACER-RARL and RACER-SAC MixedNE-LD, by incorporating the key components of RACER into the original algorithms without altering their default hyperparameters (as shown in Figure \ref{scalability}). The experimental results demonstrate consistent and stable improvements across different algorithmic instantiations. Compared to their respective baselines, both RACER-RARL and RACER-SAC MixedNE-LD achieve higher returns and exhibit stronger robustness. These findings suggest that RACER serves as a general and plug-and-play enhancement mechanism that can systematically boost the performance of SAC-based robust adversarial reinforcement learning methods without requiring algorithm-specific modifications.

\section{Conclusion}

In this work, we present RACER, a novel robust adversarial RL approach that advances the field through two core contributions: (i) We introduce a risk-sensitive adversarial objective that adaptively adjusts perturbation strength based on state-dependent risk, producing more structured and informative training dynamics. (ii) We further propose a critic consistency regularization mechanism that explicitly constrains divergence between Q-value estimators, mitigating instability caused by uneven adversarial effects on critics and producing more reliable value targets for policy optimization. 
Extensive experiments in standard MuJoCo continuous control benchmarks demonstrate that RACER consistently improves performance, training stability, and adversarial robustness under simulated adversarial disturbances.

\bibliographystyle{ACM-Reference-Format}
\bibliography{ref}

\appendix

\section{Experimental Details}

\subsection{Environments}
\label{environment_settings}

The MuJoCo environments used in the experiments follow those in QARL \cite{DBLP:conf/iclr/ReddiT0CD24}, which are modified versions of the standard implementations in the DeepMind Control Suite with adversarial settings applied \cite{todorov2012mujoco, tunyasuvunakool2020dm_control}. In total, four control tasks across four environments are used in this study: {\tt Walker-Run}, {\tt Hopper-Stand}, {\tt Cheetah-Run}, {\tt Cartpole-Balance}. Each environment includes two agents, the protagonist and the adversary, both of which take standard environment states as input. The adversarial actions selected and the magnitude of adversarial force vary across environments. Similar to QARL \cite{DBLP:conf/iclr/ReddiT0CD24} and RARL \cite{pinto2017robust}, the adversary is assigned an action space different from that of the protagonist in order to impose strong perturbations by altering the environmental dynamics. The magnitude of the adversarial force is environment-dependent. Specifically, in highly sensitive environments, we limit the maximum adversarial force to prevent complete collapse of the protagonist’s policy. Detailed environment parameters are listed in Table \ref{environment parameters}. It is worth noting that in CAT and MixedNE-LD, following the contrast settings in QARL \cite{DBLP:conf/iclr/ReddiT0CD24}, the protagonist and adversary share the same action space, deviating from the configurations in Table \ref{environment parameters}.

\begin{table*}[!h]
\caption{Environment-specific parameters}
\label{environment parameters}
\centering
\resizebox{0.9\hsize}{!}{
\begin{tabular}{c|ccl}
\toprule
\textbf{Environment} & \textbf{Adversary max force} & \textbf{Performance lower bound} & \textbf{Adversary action description} \\
\midrule
Cartpole & 0.005 & 10 & 2D force on pole (2) \\
Cheetah & 1.0 & 40 & 2D force on feet \& torso (6) \\
Hopper & 1.0 & 5 & 2D force on foot \& torso (4) \\
Walker & 1.0 & 10 & 2D forces on feet (4) \\
\bottomrule
\end{tabular}
}
\end{table*}

\begin{table*}[!h]
\caption{Definition of protagonist status and corresponding risk thresholds}
\label{risk_state}
\centering
\resizebox{0.65\hsize}{!}{
\begin{tabular}{c|ccc}
\toprule
\textbf{Environment} & \textbf{State variable} & \textbf{Risk measurement} & \textbf{Risk threshold} \\
\midrule
\multirow{3}{*}{Walker} 
& torso height $h$ 
& $\max(0, h_{\text{lim}} - h)$ 
& $h_{\text{lim}} = 1.2$ \\
& torso uprightness $u$ 
& $\max(0, u_{\text{lim}} - u)$ 
& $u_{\text{lim}} = 1$ \\
& velocity $v$ 
& $\max(0, |v| - v_{\text{lim}})$ 
& $v_{\text{lim}} = 2$ \\
\midrule
\multirow{3}{*}{Hopper}
& torso height $h$
& $\max(0, h_{\text{lim}} - h)$
& $h_{\text{lim}} = 0.6$ \\
& velocity $v$
& $\max(0, |v| - 2.5 \times v_{\text{hop}})$
& $v_{\text{hop}} = 2$ \\
& control risk $u$
& $\|u\|_2$
& - \\
\midrule
\multirow{3}{*}{Cheetah} 
& torso height $h$ 
& $\max(0, h_{\min} - h)$ 
& $h_{\min} = 0.5$ \\
& torso pitch $\theta$
& $\max(0, |\theta| - \theta_{\max})$ 
& $\theta_{\max} = 0.5$ \\
& velocity $v$ 
& $\max(0, |v - v_{\text{run}}| - \delta_v)$ 
& $\delta_v = 5.0$ \\
\midrule
\multirow{3}{*}{Cartpole}
& pole angle $\theta$
& $\max(0, |\theta| - 0.5 \times \theta_{\text{lim}})$
& $\theta_{\text{lim}} = 0.2$ \\
& angular velocity $\alpha$
& $\max(0, |\alpha| - \alpha_{\text{lim}})$
& $\alpha_{\text{lim}} = 5.0$ \\
& control risk $u$
& $\|u\|_2$
& - \\
\bottomrule
\end{tabular}
}
\end{table*}

\begin{figure*}[!t]
\centering
\includegraphics[width=.95\linewidth]{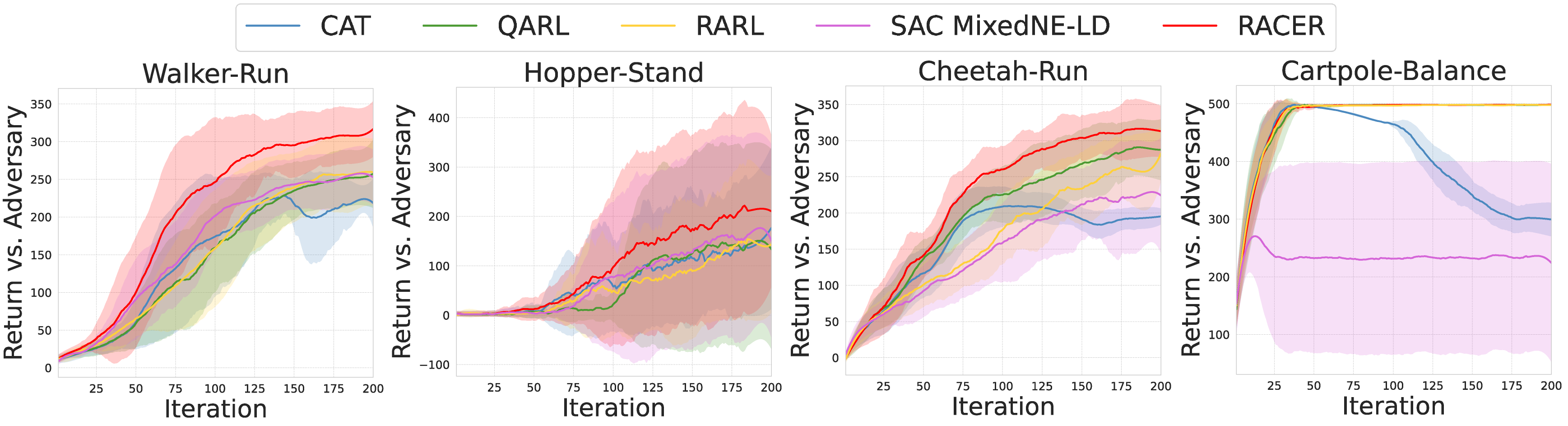} 
\caption{Adversarial performance evaluation curves comparing RACER against five baselines across four MuJoCo control and locomotion tasks. Curves show mean robustness performance over 5 random seeds with shaded regions indicating standard deviation.}
\label{vs_adversary_process}
\end{figure*}

Apart from the basic environment setup, the risk-sensitive adversarial formulation of RACER also incorporates an explicit assessment of the protagonist’s risk status. Specifically, Table \ref{risk_state} records the risk thresholds for the protagonist in each environment, along with the corresponding state dimensions. A state value exceeding its associated threshold indicates that the protagonist is at risk of losing control in that dimension; moreover, the magnitude of the deviation from the threshold serves as a quantitative indicator of risk severity.

\subsection{Hyperparameter Setting}
\label{hyperparameter_settings}

\begin{table*}[!ht]
\caption{Algorithm hyperparameters}
\label{agent and algorithm}
\centering
\resizebox{0.6\hsize}{!}{
\begin{tabular}{ll}
\toprule
\textbf{Hyperparameter} & \textbf{Value} \\
\midrule
\multicolumn{2}{l}{\textit{Shared (ALL)}} \\
\quad number of iterations ($N$) & 200 \\
\quad number of episodes per agent per iteration & 5 \\
\quad number of evaluation rollouts per iteration & 10 \\
\quad number of hidden layers & 3 \\
\quad number of hidden units per layer & 256 \\
\quad discount factor & 0.99 \\
\quad horizon & 500 \\
\midrule
\multicolumn{2}{l}{\textit{RACER}} \\
\quad consistency regularization coefficient ($\lambda$) & $1 \times 10^{-4}$\\
\midrule
\multicolumn{2}{l}{\textit{Shared (SAC and SAC MixedNE-LD)}} \\
\quad nonlinearity & ReLU \\
\quad critic optimiser & Adam \\
\quad critic learning rate & $3 \times 10^{-4}$ \\
\quad actor learning rate & $1 \times 10^{-4}$ \\
\quad initial replay memory size & $3 \times 10^{3}$ \\
\quad max replay memory size & $1 \times 10^{6}$ \\
\quad warmup transitions & $5 \times 10^{3}$ \\
\quad batch size & 256 \\
\quad target smoothing coefficient ($\tau$) & $5 \times 10^{-3}$ \\
\quad target update interval & 1 \\
\quad policy log std bounds & $[-20, 2]$ \\
\quad initial temperature & $5 \times 10^{-3}$ \\
\quad temperature learning rate & $3 \times 10^{-4}$ \\
\quad target entropy & $-\textrm{dim}(\mathcal{A})$ \\
\midrule
\multicolumn{2}{l}{\textit{SAC}} \\
\quad actor optimiser & Adam \\
\midrule
\multicolumn{2}{l}{\textit{SAC MixedNE-LD}} \\
\quad adversary influence ($\sigma$) & 0.1 \\
\quad actor optimiser & SGLD \\
\quad thermal noise ($\sigma_t$) & $10^{-3} \times (1 - 5 \times 10^{-5})^t$ \\
\quad RMSProp parameter ($\alpha$) & 0.999 \\
\quad RMSProp parameter ($\epsilon$) & $10^{-8}$ \\
\midrule
\multicolumn{2}{l}{\textit{CAT}} \\
\quad curriculum start iteration & $0.2 \times N$ \\
\quad curriculum end iteration & $0.8 \times N$ \\
\multicolumn{2}{l}{\textit{CAT Adversary}} \\
\quad gradient descent learning rate & 3 \\
\quad gradient descent step limit & 25 \\
\quad gradient descent convergence threshold ($\epsilon$) & $10^{-3}$ \\
\quad disturbance ($L^p$-norm) & 2 \\
\midrule
\multicolumn{2}{l}{\textit{QARL}} \\
\quad initial gamma distribution concentration ($k_{\text{initial}}$) & 50 \\
\quad target gamma distribution concentration ($k_{\text{target}}$) & 1 \\
\quad fixed gamma distribution rate & 1000 \\
\quad $D_{\text{KL}}$ constraint ($\epsilon$) & 0.5 \\
\quad number of rollouts needed for update & 30 \\
\bottomrule
\end{tabular}
}
\end{table*}

Table \ref{agent and algorithm} presents the hyperparameters used by RACER and each of the baseline methods. To ensure the fairness and validity of the comparisons, we strictly follow the experimental settings proposed by QARL \cite{DBLP:conf/iclr/ReddiT0CD24}, maintaining consistent parameter configurations for all baseline methods. This approach helps attribute performance differences to the methods themselves rather than to variations in hyperparameter tuning. Except for the adversary agent in the CAT method, all agents are based on SAC or its derivative variants. This design enables seamless integration of the RACER approach into existing baseline algorithms without altering their core structures, thereby allowing for a comprehensive evaluation of RACER’s generality and effectiveness.

\subsection{Hardware and Software}
\label{hardware}

All experiments were conducted on a TITAN Xp (12GB) system with 15GB of memory and an Intel(R) Xeon(R) CPU E5-2680 v4. The algorithms are based on the open-source implementation provided by QARL \cite{DBLP:conf/iclr/ReddiT0CD24}, which is built using the MushroomRL library \cite{d2021mushroomrl}. This library is also used to implement the agents and the adversarial environment wrappers.

\section{Additional Experimental Results}

\subsection{Competing Process against Adversary}

In this section, we further supplement the results for the four MuJoCo tasks mentioned in the main text ({\tt Walker-Run}, {\tt Hopper-Stand}, {\tt Cheetah-Run}, and {\tt Cartpole-Balance}) by presenting the return curves of the protagonist agent under adversarial training. As shown in Figure \ref{vs_adversary_process}, during training, the strength of adversarial perturbations increases as the adversary's policy gradually improves. Nevertheless, the protagonist controlled by RACER consistently maintains a stable and steadily increasing return, demonstrating both rapid and robust performance gains. In contrast, other baseline methods often exhibit performance fluctuations or slower return improvements when facing increasingly stronger adversaries. These results highlight RACER’s strong adaptability and robustness in dynamic and complex adversarial settings, effectively mitigating the negative impact of adversarial perturbations on the protagonist's policy and enabling more stable and efficient policy learning in adversarial environments.

\section{Ethical Considerations}

This work develops adversarial robustness techniques for reinforcement learning that could, in principle, be misused to attack deployed AI systems. We caution against such misuse and emphasize that all experiments were conducted in controlled simulation environments without real-world safety-critical deployment. The authors report all results honestly, including underperforming cases, and advocate for governance frameworks to ensure these techniques serve societal benefit.

\section{Broader Impact}

RACER takes a step toward more stable and reliable reinforcement learning under adversarial and uncertain conditions, which is particularly relevant for safety-critical applications such as robotics, autonomous systems, and control in dynamic environments. By shifting adversarial training from purely worst-case disturbances to risk-aware and state-adaptive perturbations, the framework encourages informationally useful rather than destructive interactions, potentially reducing failure modes during training and deployment. This may contribute to safer policy learning and improved robustness when agents operate in the presence of disturbances or unexpected events. However, the same principles could also be misused. More effective adversarial training techniques may be applied to design stronger attack strategies against learning systems, especially if the risk modeling is inverted or exploited to identify system vulnerabilities.  

From a societal perspective, improving robustness in learning systems is broadly beneficial, but care must be taken to ensure that such systems are validated under realistic conditions and do not create a false sense of safety. Future work on RACER should therefore emphasize transparent risk modeling, rigorous real-world evaluation, and alignment with safety standards, particularly in human-interacting domains.

\end{document}